\ifdefined\pdfoutput\pdfoutput=1\fi 
\documentclass{bvm} 

\usepackage{comment}
\hypersetup{
  colorlinks=true,
  linkcolor=blue,
  citecolor=blue,
  urlcolor=blue,
}

\begin{document}

\selectlanguage{english} 

\title{How Reliable Is Individualized Functional-Network Attribution in Resting-State fMRI Cognitive Prediction in ADNI?}
\subtitle{A Scan--Rescan Evaluation}
\titlerunning{Reliability of Individualized rs-fMRI Attribution}


\author{
	\fname{Sheethal} \lname{Bhat} \inst{1}
	\affiliation{Pattern Recognition Lab, Friedrich-Alexander-Universit\"at Erlangen-N\"urnberg (FAU)}
	\authorsEmail{sheethal.bhat@fau.de}
	\street{Martensstr.} \housenumber{3} \zipcode{91058} \city{Erlangen} \country{Germany}
	\isResponsibleAuthor,
	\fname{Mahfuzur Rahman} \lname{Chowdhury} \inst{1}
	\affiliation{Pattern Recognition Lab, Friedrich-Alexander-Universit\"at Erlangen-N\"urnberg (FAU)}
	\street{Martensstr.} \housenumber{3} \zipcode{91058} \city{Erlangen} \country{Germany},
	\fname{Imam Ahmed} \lname{Shaik} \inst{2}
	\affiliation{Siemens Healthineers}
	\authorsEmail{imamahmed.shaik@siemens-healthineers.com}
	\street{Henkestr.} \housenumber{127} \zipcode{91052} \city{Erlangen} \country{Germany}
	,
	\fname{Rose Dawn} \lname{Bharat} \inst{1}
	\affiliation{Pattern Recognition Lab, Friedrich-Alexander-Universit\"at Erlangen-N\"urnberg (FAU)}
	\authorsEmail{drrosedawnbharat@gmail.com}
	\street{Martensstr.} \housenumber{3} \zipcode{91058} \city{Erlangen} \country{Germany},
	\fname{Siming} \lname{Bayer} \inst{1}
	\affiliation{Pattern Recognition Lab, Friedrich-Alexander-Universit\"at Erlangen-N\"urnberg (FAU)}
	\street{Martensstr.} \housenumber{3} \zipcode{91058} \city{Erlangen} \country{Germany},
	\fname{Andreas} \lname{Maier} \inst{1}
	\affiliation{Pattern Recognition Lab, Friedrich-Alexander-Universit\"at Erlangen-N\"urnberg (FAU)}
	\authorsEmail{andreas.maier@fau.de}
	\street{Martensstr.} \housenumber{3} \zipcode{91058} \city{Erlangen} \country{Germany}
}
\authorrunning{Bhat et al.}

\institute{
$^1$Pattern Recognition Lab, Friedrich-Alexander-Universit\"at Erlangen-N\"urnberg (FAU), Martensstr.~3, 91058 Erlangen, Germany\\
$^2$Siemens Healthineers, Henkestr.~127, 91052 Erlangen, Germany
}

\email{sheethal.bhat@fau.de}

\maketitle

\begin{abstract}
Individualized attribution estimates which functional networks drive a model prediction for a given subject and is often proposed as an interpretable resting-state fMRI (rs-fMRI) biomarker, but its test--retest reliability is rarely quantified.
We evaluate per-subject network-ablation attribution for a frozen resting-state functional connectivity (FC) predictor (Schaefer-200 FC predicting MMSE/ADAS-Cog13) trained on $N{=}875$ subjects from the Alzheimer's Disease Neuroimaging Initiative (ADNI), using (i) within-session split-half, (ii) scan--rescan repeat visits (74 subjects, 205 baseline–repeat pairs; median inter-visit interval 24.4 months), and (iii) a subject-level permutation null.
Split-half reliability is near-ceiling (Spearman $\rho{=}0.948$/$0.950$) but drops on scan--rescan ($\rho{=}0.609$/$0.616$).
Under the mismatched-subject null, correlations remain substantial ($\rho{\approx}0.46$--$0.47$), indicating strong population-shared structure, yet matched pairs exceed this null ($z{\approx}6.6$--$6.9$, $p\le 10^{-4}$).
Thus, in ADNI, individualized network attribution contains a modest but non-zero person-specific component atop a strong population-shared pattern. We recommend reporting scan--rescan reliability together with a permutation-null control, rather than within-session proxies alone, to avoid overstating individual specificity.
\end{abstract}

\section{Introduction}

Models that map resting-state functional connectivity (FC) to cognition and clinical phenotypes are increasingly used in Alzheimer's disease (AD) neuroimaging~\cite{dadi2019benchmarking,svaldi2021identifiability}. Beyond group-level accuracy, a common objective is \emph{individualized attribution}: for a given subject, estimating which functional networks drive that subject's prediction. Such per-subject attribution is only meaningful if it is reproducible.

Reliability is a central constraint. Large-scale analyses suggest that stable brain--behavior effects often require much larger samples than historically used~\cite{marek2022reproducible}, and FC test--retest reliability is frequently modest (pooled mean edge-level Intraclass Correlation Coefficient, ICC $\approx 0.29$)~\cite{noble2019influences}. 
The ICC describes how reproducibly a single connectivity edge is measured. Prior work on subject-specific parcellations reports reproducibility across short-interval repeat sessions~\cite{wang2015parcellation,kong2019individual}, but does not quantify test--retest reliability of model-derived attribution signatures over the multi-year intervals typical of aging cohorts.

An additional aspect is \emph{attribution stability}: feature-importance maps can
vary substantially under resampling, noise, and other perturbations even when
predictive accuracy is similar~\cite{adebayo2018sanity,ghorbani2019fragile}. In
FC prediction, however, attribution work has largely emphasized group-averaged
importance. For connectomes rather than attributions, differential
identifiability formalizes the person-specific, test--retest-identifiable
component~\cite{amico2018identifiability}, and permutation-null models are
standard for separating population-shared from person-specific
structure~\cite{finn2015functional,nichols2002permutation}. Neither has been
applied to per-subject attribution signatures. To the best of our knowledge, no prior work
combines per-subject attribution of an FC-based cognitive predictor with a
quantitative test--retest evaluation on independent repeat scans of the same
individuals.

We address this gap using a frozen FC-based multilayer perceptron (FC-MLP) trained to jointly predict MMSE and ADAS-Cog13 from Schaefer-200 FC on $N{=}875$ ADNI subjects (Sect.~\ref{sec:methods}). Instead of edge-level ICC, we assess test--retest consistency of each subject's network-attribution ranking. We compute per-subject network-ablation signatures and evaluate reproducibility with three checks: within-session split-half, scan--rescan across independent repeat visits, and a cluster-aware subject-level permutation null.
The main contributions in this study are:
\begin{enumerate}
  \item Quantify how split-half overestimates scan--rescan reliability of per-subject network attribution (74 subjects; 205 session pairs).
  \item Use a subject-level permutation null to estimate the mismatched-subject baseline and the residual person-specific component.
  \item Provide reporting recommendations to avoid overstating individual specificity.
\end{enumerate}

\begin{figure}[t]
\centering
\includegraphics[width=0.92\linewidth]{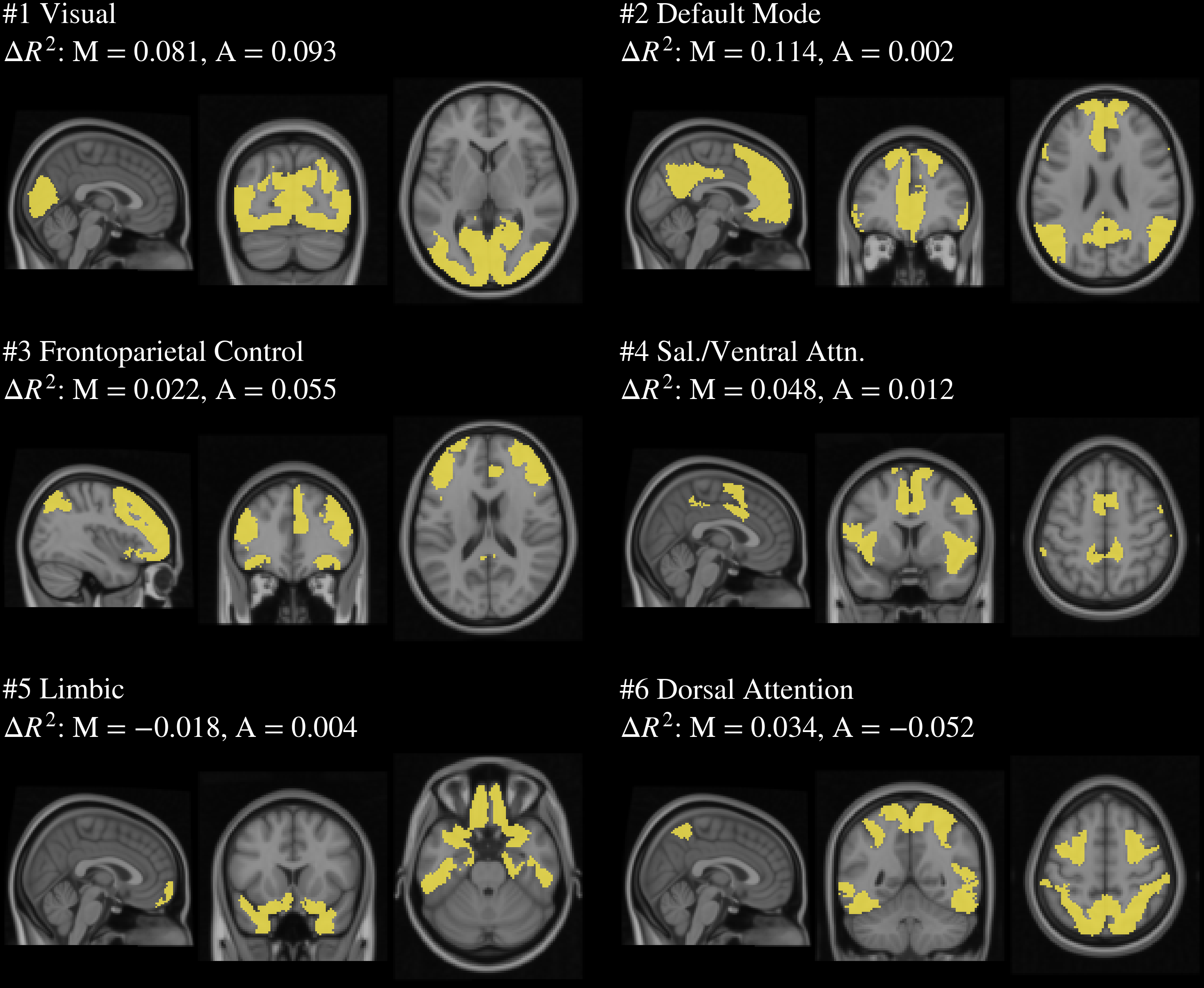}
\caption{Top 6 of the 7 Yeo networks by group-level importance, visualized on anatomical MRI. Group-level importance is quantified by the drop in test-set explained variance when ablating each network, $\Delta R^2 = R^2 - R^2_{\setminus n}$, computed on the held-out test set and averaged over three independently trained runs (shown separately for MMSE and ADAS-Cog13). $\Delta R^2$ can exceed the total $R^2$ when $R^2_{\setminus n}<0$. The displayed ``combined MMSE+ADAS-Cog13 rank'' orders networks by the summed $\Delta R^2$ across the two targets. The 7th (lowest-ranked, Somatomotor) network is omitted for layout only; all 7 are used in every analysis.}
\altText{Six anatomical brain renderings, one per Yeo functional network (Visual, Dorsal Attention, Salience/Ventral Attention, Limbic, Frontoparietal Control, Default Mode), each labeled with its group-level importance rank and importance value for MMSE and ADAS-Cog13.}
\label{fig:netimportance}
\end{figure}

\section{Methods}\label{sec:methods}

To attribute a prediction to functional networks for a single subject, let $f$ denote the frozen
FC-MLP mapping a subject's connectivity vector $x$ to a two-dimensional prediction
$\hat{\mathbf{y}}=f(x)$ with targets $t\in\{\text{MMSE},\text{ADAS-Cog13}\}$. For a given target $t$,
we write the scalar prediction as $\hat{y}^{(t)}=[f(x)]_t$.

For each of the seven Yeo networks~\cite{yeo2011organization} $n \in \{$Vis, SomMot, DorsAttn, \mbox{SalVentAttn}, Limbic, Cont, Default$\}$ (see Fig.~\ref{fig:netimportance}), we ablate the input by zeroing all FC edges incident to nodes in $n$. Because the model input concatenates an FC vector and an edge-coverage mask, $x=[\mathrm{FC};\mathrm{mask}]$, we apply the ablation to both operands: the ablated edges are set to zero in the FC vector and their corresponding mask entries are also set to zero. This yields $x_{\setminus n}$ and $\hat{y}^{(t)}_n=[f(x_{\setminus n})]_t$. The per-target network importance is then
$I^{(t)}_n=\hat{y}^{(t)}-\hat{y}^{(t)}_n$, and the subject-specific importance signature for target $t$ is
$\mathbf{I}^{(t)}=(I^{(t)}_1,\ldots,I^{(t)}_7)$.

This implementation reproduces the group-level ablation results (Fig.~\ref{fig:netimportance}) for all 7 × 2 network–target combinations to six decimal places. 
Attribution reliability between two occasions $a,b$ for the same subject is the Spearman rank correlation
$\rho(\mathbf{I}_a,\mathbf{I}_b)$, i.e., the agreement in the within-subject network ranking.

\begin{figure}[t]
  \centering
  \includegraphics[width=\textwidth]{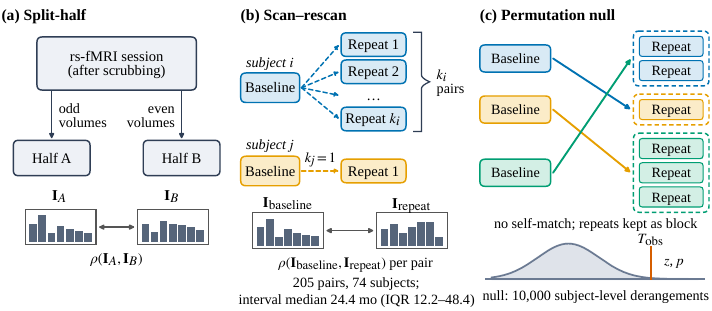}
  \caption{Overview of the three progressively stricter reliability checks. (a) Within-session split-half: odd and even volumes of one session are each attributed and compared by Spearman $\rho$. (b) Scan--rescan: each subject's baseline is compared with all repeat visits (median 24.4 months apart). (c) Permutation null: 10{,}000 subject-level derangements match each baseline with another subject's repeat visits (kept as a block, no self-match); the observed mean $T_{\mathrm{obs}}$ is compared with this null.}
  \altText{Three-panel schematic. Left: one rs-fMRI session is split into odd and even volumes, each yielding a seven-network importance signature, and the two signatures are correlated. Middle: a subject's baseline scan is paired with each of their repeat visits and each pair's signatures are correlated. Right: baselines are paired with other subjects' repeat visits to form a null distribution, against which the observed mean correlation is compared.}
  \label{fig:schematic}
\end{figure}

Fig.~\ref{fig:schematic} summarizes three reliability checks. (i) {Split-half:} for each test subject ($n{=}132$), we split the denoised rs-fMRI time series after scrubbing into odd/even surviving volumes, recompute FC and attribution per half, and compute $\rho(\mathbf{I}_{\text{half-A}},\mathbf{I}_{\text{half-B}})$ (minimum 30 volumes per half; 0 exclusions). We additionally test whether split-half $\rho$ increases with per-half scan duration. (ii) {Scan--rescan:} we attribute independent repeat-visit rs-fMRI and compute $\rho(\mathbf{I}_{\text{baseline}},\mathbf{I}_{\text{repeat}})$ for baseline $\times$ all follow-ups, yielding 205 pairs from 74 subjects with uneven repeats ($k_i\in[1,12]$; 30 subjects contribute 1 pair, 18 contribute 2, 13 contribute 3; tail to 12). 
(iii) Permutation null: for $B{=}10{,}000$ permutations, we draw a derangement $\pi$ of the 74 subjects, pair each subject $i$'s baseline signature with the repeat signatures of subject $\pi(i)$ (repeats kept as a block), and recompute the subject-level mean reliability. With $T_{\mathrm{obs}}$ the observed subject-level mean and $b$ the number of permutations with statistic $\ge T_{\mathrm{obs}}$, we report $p=(b+1)/(B+1)$ and $z=(T_{\mathrm{obs}}-\mu_{\mathrm{null}})/\sigma_{\mathrm{null}}$~\cite{nichols2002permutation}.


\paragraph{Artifact checks} {Edge-count:} let $c\in\mathbb{R}^7$ be the fixed vector of ablated edge counts (unique edges zeroed when ablating each network). For each subject we compute the across-network Spearman correlation $\rho(\mathbf{I}^{(t)},c)$ (7 items) and summarize it across subjects by its mean, separately per target; with only 7 items, Spearman $\rho$ is highly discrete, so this check is treated as descriptive. For scan--rescan, we first average each subject's per-pair $\rho(\mathbf{I}^{(t)},c)$ across that subject's repeat visits and then compute the across-subject mean.
{Scan duration:} we correlate (Spearman, across the 132 test subjects) each subject's per-half scan duration (seconds; half~A, equivalent to half~B up to one volume) with split-half $\rho$.


\paragraph{Aggregation and inference} All Spearman correlations are computed separately within each of the three trained runs and averaged across runs. 
Spearman $\rho$ uses average ranks for ties; no ties occurred in any importance signature (verified for all 1,674 scan--rescan signatures, i.e., 279 sessions $\times$ 2 targets $\times$ 3 runs, and all split-half signatures).
As subjects contribute unequal numbers of repeat pairs, scan--rescan is summarized at the subject level by first averaging each subject's per-pair $\rho$ values and then summarizing across subjects; we also report the pair-pooled mean for comparability. Uncertainty on headline scan--rescan and split-half means is a 95\,\% percentile bootstrap CI over 10{,}000 resamples of subjects (resampling whole subjects). We test the split-half versus scan--rescan gap with a Wilcoxon signed-rank test across the 74 subjects present in both analyses.

\begin{table}[ht]
\centering
\caption{Attribution reliability (Spearman $\rho$ of the per-subject 7-network importance signature). The subject is the unit of analysis (split-half: $n{=}132$; scan--rescan: $n{=}74$, with pairs averaged within subject). CIs are subject-level bootstrap intervals; ``run SD'' is the SD across 3 runs; ``--'' indicates not computed. Permutation-null rows use 10{,}000 subject-level derangements.}
\label{tab:reliability}
\small
\begin{tabular*}{\linewidth}{@{\extracolsep{\fill}}lcccc@{}}
\hline
 & \multicolumn{2}{c}{Split-half} & \multicolumn{2}{c}{Scan--rescan} \\
Statistic & MMSE & ADAS-Cog13 & MMSE & ADAS-Cog13 \\
\hline
Mean               & 0.948 & 0.950 & 0.609 & 0.616 \\
95\,\% CI          & 0.929--0.964 & 0.934--0.964 & 0.553--0.663 & 0.571--0.660 \\
Median             & 0.988 & 0.988 & 0.625 & 0.661 \\
Std                & 0.101 & 0.085 & 0.242 & 0.198 \\
10--90\% & 0.87--1.00 & 0.85--1.00 & 0.34--0.90 & 0.33--0.85 \\
Run SD ($n{=}3$)   & 0.005 & 0.007 & 0.033 & 0.020 \\
\hline
Null mean $\pm$ SD & \multicolumn{2}{c}{--} & $0.463\pm0.021$ & $0.472\pm0.022$ \\
Null range         & \multicolumn{2}{c}{--} & 0.379--0.549 & 0.393--0.562 \\
$z$-score          & \multicolumn{2}{c}{--} & 6.89 & 6.63 \\
$p$-value          & \multicolumn{2}{c}{--} & $\le 10^{-4}$ & $\le 10^{-4}$ \\
\hline
Pair-pooled mean   & \multicolumn{2}{c}{--} & 0.593 & 0.588 \\
\hline
\end{tabular*}
\end{table}
\section{Experimental details}
\label{results}

\subsection{Data description}
ADNI is a multi-site, longitudinal study of aging and Alzheimer's disease with multimodal neuroimaging~\cite{jack2008alzheimer}. We analyzed data from $N{=}875$ unique ADNI participants who had \emph{both} T1-weighted MRI and rs-fMRI scans. 
Within the $N{=}875$ cohort, diagnoses are CN ($n{=}464$), MCI ($n{=}315$), Dementia ($n{=}94$), and unknown ($n{=}2$). Age is $72.6\pm7.9$ years (range 50--97); 489 subjects are female and 386 male. Subjects were enrolled in ADNI1 ($n{=}37$), ADNI-GO ($n{=}38$), ADNI2 ($n{=}256$), ADNI3 ($n{=}474$), and ADNI4 ($n{=}68$); 2 unknown. rs-fMRI scans were acquired at 3.0~T across 53 sites with heterogeneous protocols (TR 0.607--6.02~s; 96--972 volumes after removal of initial volumes).
For the upstream prediction model, the cohort is split once at the subject level into fixed train/validation/test partitions of 611/132/132 stratified by diagnosis. We evaluate attribution and reliability only on the held-out 132-subject test split; scan--rescan further restricts to test subjects with repeat-visit rs-fMRI.

\subsection{Implementation details}
We analyze a trained FC-MLP and keep its weights fixed for all attribution and reliability analyses.

\paragraph{Input features} rs-fMRI preprocessing includes motion correction, HD-BET brain extraction~\cite{XXXX-isensee2019}, rigid rs-fMRI$\to$T1 registration, detrending, and nuisance regression of mean white-matter (WM) and cerebrospinal-fluid (CSF) signals plus 5 anatomical CompCor (aCompCor) components~\cite{XXXX-behzadi2007} from a combined WM/CSF mask, without global signal regression. Time series are band-pass filtered (0.01--0.1\,Hz, using each subject's TR), $z$-scored, and scrubbed at framewise displacement (FD) $>0.5$\,mm after filtering; FD was computed from translation magnitude only rather than the full six-parameter formulation~\cite{power2012spurious}, as rotation parameters were not retained upstream. No spatial smoothing or slice-timing correction is applied. We compute Schaefer-200~\cite{schaefer2018local} FC as a Fisher-$z$ transformed $200\times200$ Pearson correlation matrix and vectorize the upper triangle (19,900 edges). An edge-coverage mask is concatenated, yielding a 39,800-dimensional input.

\paragraph{Model and training (frozen)} The FC-MLP encoder maps $39{,}800\to256\to128$ with LayerNorm/GELU/dropout (0.3), and the decoder maps $128\to64\to2$ with GELU/dropout (0.3). The model is trained with Huber loss ($\delta{=}1.0$) using Adam ($\mathrm{lr}=10^{-4}$, weight decay $10^{-5}$; no schedule/warmup), batch size 16, and up to 100 epochs. Early stopping uses patience 15 on validation loss; we keep the single best checkpoint (lowest validation loss). During training only, an augmentation zeros 5 randomly chosen parcels per sample. Targets are $z$-scored on train only and inverse-transformed for reporting. Three independent runs with different random initializations share the same fixed split. ComBat harmonization~\cite{XXXX-fortin2017} uses the TR group as batch variable (a proxy for acquisition protocol) and protects diagnosis, whose composition differs across TR groups ($\chi^2{=}35.66$, $p<0.001$); no other covariates are included. It is fit on the training split only, never refit, and applied to validation/test subjects and split-half halves, but not to repeat-visit sessions (both sides of scan--rescan use pre-ComBat features).

\subsection{Results}
The frozen predictor is weak: on the $n{=}113$ test subjects with both targets, $R^2{=}0.053\pm0.034$ for MMSE and $0.087\pm0.037$ for ADAS-Cog13 (mean $\pm$ SD over runs). As a diagnostic check, the predicted ADAS-Cog13 score separates CN from Dementia with AUC $0.780$ (95~\% bootstrap CI 0.611–0.911; $n=72$ of the 113 subjects with both targets: 58 CN, 14 Dementia).

Within-session split-half reliability is near-ceiling for both targets (mean $\rho{=}0.948$/$0.950$, median $\approx0.99$; Tab.~\ref{tab:reliability}). On independent repeat visits (205 session pairs from 74 subjects), subject-level scan--rescan reliability drops to $\rho{=}0.609$ (MMSE) and $0.616$ (ADAS-Cog13), with large between-subject variation (10th--90th percentiles $0.34$--$0.90$ and $0.33$--$0.85$). Pair-pooled means are slightly lower ($0.593$/$0.588$) because subjects with more follow-ups receive more weight. Restricted to the same 74 subjects, split-half reliability is $0.933$/$0.939$, a paired difference of $0.324$/$0.323$ (Wilcoxon $p<10^{-12}$), indicating that split-half substantially overestimates test--retest reliability.

Under the subject-level permutation null, mismatched subjects still correlate strongly ($\rho{=}0.463$/$0.472$), but the matched statistic exceeds the null ($z{=}6.89$/$6.63$; $p\le10^{-4}$), corresponding to a modest person-specific excess of $\approx0.15$ (MMSE) and $\approx0.14$ (ADAS-Cog13).
The artifact checks do not indicate a network-size effect: mean edge-count correlations $\rho(\mathbf{I}^{(t)},c)$ are weakly negative for split-half (MMSE $-0.120$; ADAS-Cog13 $-0.163$) and repeat visits (MMSE $-0.112$; ADAS-Cog13 $-0.158$). Split-half reliability increases with per-half scan duration ($\rho{=}0.199$/$0.277$), consistent with sampling-noise limits, although duration is confounded with acquisition protocol in ADNI.

\section{Discussion}

Per-subject attribution reliability depends strongly on protocol: split-half is near-ceiling ($\rho{\approx}0.95$), but drops on scan--rescan across independent repeat visits ($\rho{\approx}0.61$--$0.62$), consistent with additional between-visit variability beyond within-session sampling noise. The permutation null suggests that much of this structure is population-shared, with only a modest person-specific excess; robustness to other cohorts, preprocessing, parcellations, attribution operators, and models remains untested.

The coexistence of shared and person-specific structure mirrors FC itself, where connectomes are identifiable across sessions~\cite{finn2015functional,amico2018identifiability} yet test--retest reliability is modest~\cite{noble2019influences}. The strong shared profile likely has two sources. First, the ablations overlap by construction: every between-network edge $(i,j)$ with $i\in n$, $j\in m$ is zeroed in both $x_{\setminus n}$ and $x_{\setminus m}$, coupling $I^{(t)}_n$ and $I^{(t)}_m$; within-network-only ablation would remove this overlap. Second, the weak predictor may respond mainly to group-level FC structure. Because $\rho$ is bounded above, a high shared baseline leaves little room for an individual excess, so the person-specific component may be underestimated.

Limitations include the modest repeat-visit sample (74 subjects; 205 session pairs) and the weak, run-variable predictor (Sect.~\ref{results}). Repeat-visit sessions were not ComBat-harmonized, so part of the split-half to scan--rescan gap may reflect this mismatch. Inter-visit intervals are long and variable (median 24.4 months, IQR 12.2--48.4) and are not modeled.

We therefore recommend treating within-session proxies as expected upper bounds, reporting scan--rescan estimates when repeat visits exist, and including a permutation-null (or equivalent) control; these recommendations should be validated in further aging/AD cohorts such as OASIS-3 and AIBL. In summary, per-subject network attributions are far less reproducible across visits than within-session estimates suggest, and most of their apparent stability reflects a pattern shared across the cohort, so individual specificity should be demonstrated against an explicit population baseline before attributions are interpreted as personal biomarkers.

\begin{acknowledgement}
We acknowledge HPC support from XXXXXX and partial hardware funding from the XXXXX.
Data were obtained from the Alzheimer's Disease Neuroimaging Initiative (ADNI; \texttt{adni.loni.usc.edu}). As such, the investigators within the ADNI contributed to the design of ADNI but did not participate in this report. 
Code, derived features, and the subject-level split file are available upon reasonable request, subject to ADNI DUA constraints.
\end{acknowledgement}

\printbibliography

@article{nichols2002permutation,
  author  = {Nichols, Thomas E. and Holmes, Andrew P.},
  title   = {Nonparametric permutation tests for functional neuroimaging: a primer with examples},
  journal = {Human Brain Mapping},
  volume  = {15},
  number  = {1},
  pages   = {1--25},
  year    = {2002},
  doi     = {10.1002/hbm.1058}
}

@article{dadi2019benchmarking,
  author  = {Dadi, Kamalaker and Rahim, Mehdi and Abraham, Alexandre and others},
  title   = {Benchmarking functional connectome-based predictive models for resting-state {fMRI}},
  journal = {NeuroImage},
  volume  = {192},
  pages   = {115--134},
  year    = {2019},
  doi     = {10.1016/j.neuroimage.2019.02.062}
}

@article{svaldi2021identifiability,
  author  = {Svaldi, Dario O. and Joshi, Ashish A. and Shattuck, David W. and others},
  title   = {Optimizing differential identifiability improves connectome predictive modeling of cognitive deficits from functional connectivity in {Alzheimer}'s disease},
  journal = {Human Brain Mapping},
  volume  = {42},
  number  = {11},
  pages   = {3500--3516},
  year    = {2021},
  doi     = {10.1002/hbm.25448}
}

@article{amico2018identifiability,
  author  = {Amico, Enrico and Go\~{n}i, Joaqu\'in},
  title   = {The quest for identifiability in human functional connectomes},
  journal = {Scientific Reports},
  volume  = {8},
  number  = {1},
  pages   = {8254},
  year    = {2018},
  doi     = {10.1038/s41598-018-25089-1}
}

@inproceedings{adebayo2018sanity,
  author    = {Adebayo, Julius and Gilmer, Justin and Muelly, Michael and Goodfellow, Ian and Hardt, Moritz and Kim, Been},
  title     = {Sanity Checks for Saliency Maps},
  booktitle = {Advances in Neural Information Processing Systems (NeurIPS)},
  year      = {2018},
  pages     = {9525--9536}
}

@inproceedings{ghorbani2019fragile,
  author    = {Ghorbani, Amirata and Abid, Abubakar and Zou, James},
  title     = {Interpretation of Neural Networks Is Fragile},
  booktitle = {Proceedings of the {AAAI} Conference on Artificial Intelligence},
  year      = {2019},
  volume    = {33},
  number    = {01},
  pages     = {3681--3688},
  doi       = {10.1609/aaai.v33i01.33013681}
}

@article{marek2022reproducible,
  author  = {Marek, Scott and Tervo-Clemmens, Brenden and Calabro, Finnegan J.
             and others},
  title   = {Reproducible brain-wide association studies require thousands
             of individuals},
  journal = {Nature},
  volume  = {603},
  pages   = {654--660},
  year    = {2022},
  doi     = {10.1038/s41586-022-04492-9}
}

@article{noble2019influences,
  author  = {Noble, Stephanie and Scheinost, Dustin and Constable, R. Todd},
  title   = {A decade of test-retest reliability of functional connectivity:
             A systematic review and meta-analysis},
  journal = {NeuroImage},
  volume  = {203},
  pages   = {116157},
  year    = {2019},
  doi     = {10.1016/j.neuroimage.2019.116157}
}

@article{wang2015parcellation,
  author  = {Wang, Danhong and Buckner, Randy L. and Fox, Michael D. and
             others},
  title   = {Parcellating cortical functional networks in individuals},
  journal = {Nature Neuroscience},
  volume  = {18},
  pages   = {1853--1860},
  year    = {2015},
  doi     = {10.1038/nn.4164}
}

@article{kong2019individual,
  author  = {Kong, Ru and Li, Jingwei and Orban, Csaba and others},
  title   = {Spatial topography of individual-specific cortical networks
             predicts human cognition, personality, and emotion},
  journal = {Cerebral Cortex},
  volume  = {29},
  number  = {6},
  pages   = {2533--2551},
  year    = {2019},
  doi     = {10.1093/cercor/bhy123}
}

@article{finn2015functional,
  author  = {Finn, Emily S. and Shen, Xilin and Scheinost, Dustin and others},
  title   = {Functional connectome fingerprinting: identifying individuals
             using patterns of brain connectivity},
  journal = {Nature Neuroscience},
  volume  = {18},
  pages   = {1664--1671},
  year    = {2015},
  doi     = {10.1038/nn.4135}
}

@article{schaefer2018local,
  author  = {Schaefer, Alexander and Kong, Ru and Gordon, Evan M. and Laumann, Timothy O. and Zuo, Xi-Nian and Holmes, Avram J. and Eickhoff, Simon B. and Yeo, B. T. Thomas},
  title   = {Local-Global Parcellation of the Human Cerebral Cortex from Intrinsic Functional Connectivity {MRI}},
  journal = {Cerebral Cortex},
  volume  = {28},
  number  = {9},
  pages   = {3095--3114},
  year    = {2018},
  doi     = {10.1093/cercor/bhx179}
}

@article{yeo2011organization,
  author  = {Yeo, B. T. Thomas and Krienen, Fenna M. and Sepulcre, Jorge and Sabuncu, Mert R. and Lashkari, Danial and Hollinshead, Marisa and Roffman, Joshua L. and Smoller, Jordan W. and Z{\"o}llei, Lilla and Polimeni, Jonathan R. and Fischl, Bruce and Liu, Hesheng and Buckner, Randy L.},
  title   = {The organization of the human cerebral cortex estimated by intrinsic functional connectivity},
  journal = {Journal of Neurophysiology},
  volume  = {106},
  number  = {3},
  pages   = {1125--1165},
  year    = {2011},
  doi     = {10.1152/jn.00338.2011}
}

@article{jack2008alzheimer,
  author  = {Jack, Clifford R. and Bernstein, Matt A. and Fox, Nick C. and Thompson, Paul and Alexander, Gene and Harvey, Danielle and Borowski, Bret and Britson, Paula J. and Whitwell, Jennifer L. and Ward, Chadwick and Dale, Anders M. and others},
  title   = {The {Alzheimer's} Disease Neuroimaging Initiative ({ADNI}): {MRI} methods},
  journal = {Journal of Magnetic Resonance Imaging},
  volume  = {27},
  number  = {4},
  pages   = {685--691},
  year    = {2008},
  doi     = {10.1002/jmri.21049}
}

@article{power2012spurious,
  author  = {Power, Jonathan D. and Barnes, Kelly A. and Snyder, Abraham Z. and Schlaggar, Bradley L. and Petersen, Steven E.},
  title   = {Spurious but systematic correlations in functional connectivity {MRI} networks arise from subject motion},
  journal = {NeuroImage},
  volume  = {59},
  number  = {3},
  pages   = {2142--2154},
  year    = {2012},
  doi     = {10.1016/j.neuroimage.2011.10.018}
}

@article{XXXX-fortin2017,
  author  = {Fortin, Jean-Philippe and Parker, Drew and Tun{\c{c}}, Birkan and Watanabe, Takanori and Elliott, Mark A. and Ruparel, Kosha and Roalf, David R. and Satterthwaite, Theodore D. and Gur, Ruben C. and Gur, Raquel E. and Schultz, Robert T. and Verma, Ragini and Shinohara, Russell T.},
  title   = {Harmonization of multi-site diffusion tensor imaging data},
  journal = {NeuroImage},
  volume  = {161},
  pages   = {149--170},
  year    = {2017},
  doi     = {10.1016/j.neuroimage.2017.08.047}
}

@article{XXXX-behzadi2007,
  author  = {Behzadi, Yashar and Restom, Khaled and Liau, Joy and Liu, Thomas T.},
  title   = {A component based noise correction method ({CompCor}) for {BOLD} and perfusion based {fMRI}},
  journal = {NeuroImage},
  volume  = {37},
  number  = {1},
  pages   = {90--101},
  year    = {2007},
  doi     = {10.1016/j.neuroimage.2007.04.042}
}

@article{XXXX-isensee2019,
  author  = {Isensee, Fabian and Schell, Marianne and Pflueger, Irada and Brugnara, Gianluca and Bonekamp, David and Neuberger, Ulf and Wick, Antje and Schlemmer, Heinz-Peter and Heiland, Sabine and Wick, Wolfgang and Bendszus, Martin and Maier-Hein, Klaus H. and Kickingereder, Philipp},
  title   = {Automated brain extraction of multisequence {MRI} using artificial neural networks},
  journal = {Human Brain Mapping},
  volume  = {40},
  number  = {17},
  pages   = {4952--4964},
  year    = {2019},
  doi     = {10.1002/hbm.24750}
}
\end{document}